# 基于YOLOv5和点云3D投影的智能驾驶车辆前方多目标跟踪检测

刘大勇　张清睿　孟泽阳

吉林大学电子科学与工程学院，吉林，长春，130012

**摘要**：在多目标跟踪检测任务中，需要连续地跟踪多个目标，如车辆、行人等。为了实现这一目标，系统必须能够连续地获取并处理包含这些目标的图像帧。这些连续帧图像使算法能够在每一帧图像中实时更新目标的位置和状态，如何准确地将检测到的目标与前一帧或后一帧中的目标进行关联，形成稳定的轨迹，是一个复杂的问题。为此，提出基于YOLOv5和点云3D投影的智能驾驶车辆前方多目标跟踪检测方法。运用Retinex算法对车辆前方环境图像进行增强处理，去除图像中光线干扰，以YOLOv5网络结构为基础搭建智能检测模型，将增强后的图像输入模型中，通过特征提取和目标定位，识别出车辆前方多目标。结合点云3D投影技术，推断相邻帧图像在投影坐标系中位置变化的关联性，将连续多帧图像的多目标识别结果依次投影到三维激光点云环境中，即可完成对车辆前方所有目标运动轨迹的有效跟踪。实验结果表明：应用该方法完成智能驾驶车辆前方多目标跟踪检测，所得结果MOTA（跟踪准确度）值大于30，证明了其优越的跟踪检测性能。

**关键词**：智能驾驶车辆；YOLOv5；多目标识别；点云3D投影；图像增强；跟踪检测

**Abstract:** In multi-target tracking and detection tasks, it is necessary to continuously track multiple targets, such as vehicles, pedestrians, etc. To achieve this goal, the system must be able to continuously acquire and process image frames containing these targets. These consecutive frame images enable the algorithm to update the position and state of the target in real-time in each frame of the image. How to accurately associate the detected target with the target in the previous or next frame to form a stable trajectory is a complex problem. Therefore, a multi object tracking and detection method for intelligent driving vehicles based on YOLOv5 and point cloud 3D projection is proposed. Using Retinex algorithm to enhance the image of the environment in front of the vehicle, remove light interference in the image, and build an intelligent detection model based on YOLOv5 network structure. The enhanced image is input into the model, and multiple targets in front of the vehicle are identified through feature extraction and target localization. By combining point cloud 3D projection technology, the correlation between the position changes of adjacent frame images in the projection coordinate system can be inferred. By sequentially projecting the multi-target recognition results of multiple consecutive frame images into the 3D laser point cloud environment, effective tracking of the motion trajectories of all targets in front of the vehicle can be achieved. The experimental results show that the application of this method for intelligent driving vehicle front multi-target tracking and detection yields a MOTA (Tracking Accuracy) value greater than 30, demonstrating its superior tracking and detection performance.

## 0 引言

智能驾驶车辆是汽车工业与人工智能、物联网等技术深度融合的产物[1]，已经成为未来交通系统的重要组成部分，与人工驾驶车辆相比能够满足现代交通系统对安全、效率与智能化水平的迫切需求[2]。但在智能驾驶系统中，多目标跟踪检测是至关重要的一环，其指的是车辆能够实时、准确地跟踪并识别前方的多个目标[3]，如行人、其他车辆、自行车等，从而实现对这些目标运动轨迹的有效预测。这一技术不仅为智能驾驶车辆提供了必要的感知能力，还是其进行决策、规划、控制等后续操作的基础[4]。因此，面向复杂多变的交通环境，探索智能驾驶车辆前方多目标跟踪检测方法，成为一项重要研究内容。

针对当前研究成果进行分析可以发现，武宏伟等[5]提出基于3D激光雷达点云处理的跟踪检测方法，面向复杂的车辆行驶场景，提出基于多层级结构的跟踪检测框架。采用基于椭圆门限的多维特征分析算法，提取环境图像中包含的特征信息，再将其导入多层级框架中，通过特征关联分析、运动状态预测和自适应滤波处理，实现对每个移动目标的有效跟踪。实验结果表明，该算法采用高效的计算流程和优化的数据结构，能够在短时间内处理大量的点云数据，使得跟踪检测结果满足实时性要求，但滤波算法在预测目标运动状态时会受到噪声和不确定性因素的影响，导致最终输出的跟踪结果MOTA值较低。张瑶等[6]提出结合短时记忆与CenterTrack的跟踪检测方法，获取包含多目标的现场环境图像后，通过小样本扩增操作，增加小目标车辆训练样本数量，提升多目标跟踪检测模型对小目标的检测能力。将短时记忆网络和CenterTrack算法（基于中心点的目标跟踪算法）结合起来，构建多目标智能跟踪检测网络模型。面向包含多目标的现场环境图像进行小样本扩增，增加小目标车辆训练样本数量，提升模型应用性

能。应用训练好的模型对现场采集数据进行分析，即可得出多目标识别与跟踪结果。通过验证可知，CenterTrack 支持多种数据集和模型，可以根据不同的场景进行定制化开发，这使得该方法具有较强的可拓展性，但 CenterTrack 算法主要依赖于目标的中心点进行跟踪，将其应用到遮挡、光照变化等复杂场景中，所得跟踪检测结果的 MOTA 值会发生大幅降低。刘毅等[7]提出基于改进 KCF 的跟踪检测方法，针对待跟踪视频序列进行分析，提取一系列图像数据，利用结合了回归思想和锚框机制的 SSD 算法，检测出图像内包含的所有目标，并将目标检测结果导入至跟踪器中，通过 KCF 算法跟踪连续图像帧检测框位置变化，得出多目标跟踪结果。实验结果表明，KCF 算法能够较好地应对复杂环境中光照不均的问题，使得多目标跟踪检测方法表现出较强的抗干扰能力，但目标尺度发生剧烈变化时，改进 KCF 算法无法及时调整跟踪窗口的大小和位置，导致跟踪结果 MOTA 值偏低。何维堃等[8]提出基于 DeepSort 的跟踪检测方法，以 YOLOX 网络结构为基础设计前端检测器，针对所有环境感知图像完成多目标检测。再结合改进的 ResNet13 网络和 SENet 网络，构成基于 DeepSort 重识别网络的多目标跟踪模型，通过对目标特征的不断跟踪，获取多目标实时变化情况。研究结果表明，DeepSort 结合了深度学习的强大表征能力与经典追踪技术，极大提升了多目标跟踪检测方法重识别能力，但 DeepSort 算法中的参数设置不当，会直接造成跟踪结果 MOTA 值降低。

考虑到智能驾驶车辆前方多目标跟踪检测的需求，提出结合 YOLOv5 和点云 3D 投影的新方法。通过 YOLOv5 算法对机器视觉 2D 图像进行分析，识别出车辆前方多目标，再将检测目标与激光雷达点云数据相融合，投影到 3D 空间内，通过观察目标 3D 位置的变化，得出多目标跟踪检测结果，为智能驾驶车辆提供可靠的支持。

# 1 智能驾驶车辆前方多目标跟踪检测方法
## 1.1 建立车辆前方环境图像增强方法

智能驾驶车辆行驶过程中，借助车载相机获取前方环境图像，是多目标跟踪检测的基础环节。考虑到环境图像采集过程中，光照条件会影响图像质量，为了得出准确的跟踪检测结果，引入 Retinex 算法对车辆前方环境图像进行增强处理[9]，去除图像中照射光线，仅保留多目标自身反射属性。在 Retinex 理论应用时，引入基于螺旋结构的迭代估计模式，对图像中的光照进行分段线性化计算，如图 1 所示。

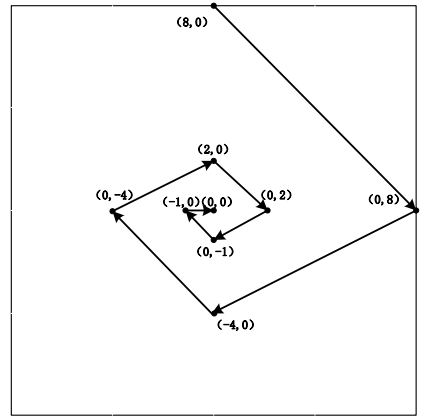

**图 1 Retinex 算法螺旋结构路径示意图**

图 1 中，将像素点（0，0）看作校正目标点，该点像素校正值是根据螺旋结构路径上拐点像素值推导出来的。为了确保像素值校正结果最优，在拐点选择分布上，应该呈现出越靠近目标点附近越密集的特点[10]。

依托于 Retinex 算法增强车辆前方环境图像时，为了达到减少计算量的目的，先将图像中目标点像素值从整数域调整到对数域[11]，具体转化过程如公式（1）所示。

$$f(i,j) = \log[a + f(i,j)] \quad (1)$$

式中，$(i,j)$ 表示图像中目标像素点坐标，$f$ 表示智能驾驶车辆前方环境图像，$\log$ 表示对数函数，$a$ 表示一个用来确保像素值不为负值的常数。比较距离目标点最远的两个螺旋结构路径拐点，并将二者之间的坐标变化量表示为：

$$D = 2^{F[L\ \min(W,H)-1]} \quad (2)$$

式中，$D$ 表示坐标变化量，$F$ 表示取整函数，$L$ 表示两个像素点之间的距离度量值，$\min$ 表示最小值取值函数，$W \times H$ 表示环境图像像素尺寸。下一步参与计算的两个拐点之间间距，仅为上一步拐点间距的 1/2，同时拐点方向也会发生顺时针旋转[12]，那么坐标变化量可以计算为：

$$D_1 = -\frac{Df(i,j)}{2} \quad (3)$$

式中，$D_1$ 表示第二步计算得出的拐点坐标变化量。

将坐标变化量代入到 Retinex 迭代计算过程中，若坐标变化量大于 0，则可以计算出图像灰度矩阵为：

$$\begin{cases} g_{t+1}(i+D,j) = g_t(i,j) + f(i+D,j) - f(i,j) \\ g_{t+1}(i,j+D) = g_t(i,j) + f(i,j+D) - f(i,j) \end{cases} \quad (4)$$

式中，$g$ 表示像素点灰度值，$t$ 表示 Retinex 迭代计算次数。另一种情况，是坐标变化量小于 0，此时公式（4）所示的计算过程将变化为公式（5）。

$$\begin{cases} g_{t+1}(i,j) = g_t(i-D,j) + f(i,j) - f(i-D,j) \\ g_{t+1}(i,j) = g_t(i,j-D) + f(i,j) - f(i,j-D) \end{cases} \quad (5)$$

通过 Retinex 迭代运算求出的像素灰度值，表示为浮点数，采用线性计算原理对其进行转化计算，即可得到有效灰度值，完成车辆前方环境图像增强处理。

$$f'(i,j) = \frac{g_{t+1}(i,j) - g_{(t+1)\min}(i,j)}{g_{(t+1)\max}(i,j) - g_{(t+1)\min}(i,j)} \times 255 \quad (6)$$

式中，$f'$ 表示增强处理后的环境图像，$\max$ 表示最大值取值函数。

### 1.2 基于 YOLOv5 的车辆前方多目标检测方法

以增强处理后的智能驾驶车辆前方环境无图像为目标，为了从图中检测出所有目标，建立包含骨干网络、颈部网络、头部网络三个模块的 YOLOv5 网络[13]模型，如图 2 所示。

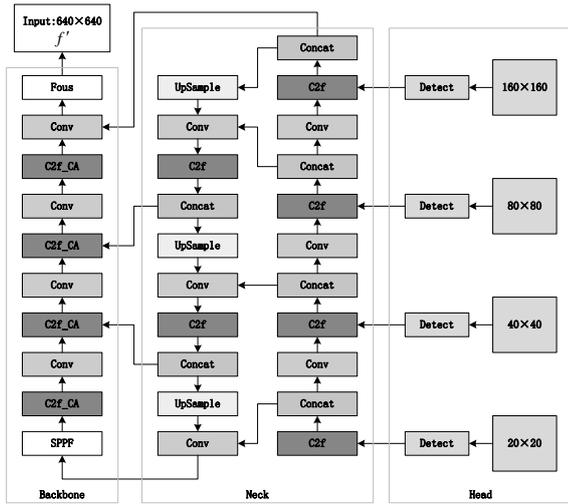

**图 2 YOLOv5 网络模型**

如图 2 所示，将增强处理后的环境图像输入 YOLOv5 网络模型中，通过卷积运算提取图像包含的特征信息，是主干网络和颈部网络两个模块计算目标。尤其是图 3（a）所示的 C2f 结构引入到 YOLOv5 网络后，其可以在通道维度上将输入图像划分为两部分，多角度学习图像包含特征[14]。

$$R_v = C_v \times f' + b_v f'(i,j) \quad (7)$$

式中，$v$ 表示 C2f 模块中卷积层编号，$R$ 表示输出特征图，$C$ 表示卷积核对应的权重系数，$b$ 表示偏置。

将图 3（b）给出的注意力机制，与 C2f 模块相结合，可以对多尺度特征提取结果进行拼接，描述图像不同特征信息之间的长距离依赖关系，凸显出图像中包含的小目标[15]。

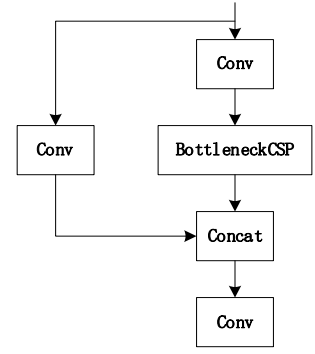

（a）C2f 结构

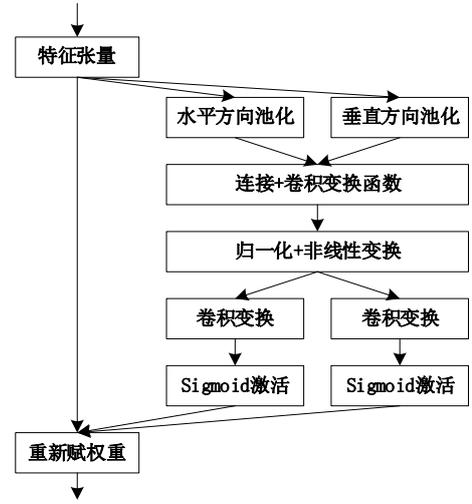

（b）坐标注意力模块

**图 3 C2f 模块和坐标注意力模块结构图**

如图 3（b）所示，C2f 模块给出结果输入到坐标注意力模块后，需要嵌入特征图中的不同特征点的坐标信息，并通过水平池化层与垂直池化层[16]，对每个通道的特征图进行编码处理。以任意一个通道特征图为例，其池化运算输出结果为：

$$\begin{cases} \beta_q(h) = \dfrac{R_v}{\hat{W}} \sum_{u=1}^{\hat{W}} R_u(h,u) \\ \beta'_q(w) = \dfrac{R_v}{\hat{H}} \sum_{p=1}^{\hat{H}} R_w(p,w) \end{cases} \quad (8)$$

式中，$\beta$ 表示水平方向池化输出结果，$\beta'$ 表示垂直方向池化输出结果，$q$ 表示通道编号，$h$、$w$ 表示目标通道输出特征图的高度和宽度，$\hat{W} \times \hat{H}$ 表示 C2f 模块输出特征图尺寸，$u$、$p$ 表示特征点横坐标和列坐标。

借助公式（8）对特征图进行池化计算，可以生成两幅独立的感知注意力特征图，将特征图连接起来，并按照公式（9）完成卷积变换和非线性变换，最终输出

包含车辆前方环境图像全局特征的中间特征图。

$$\beta'' = \delta(\varepsilon[\beta_q, \beta'_q]) \quad (9)$$

式中，$\beta''$ 表示中间特征图，$\delta$ 表示非线性激活函数，$\varepsilon$ 表示 1×1 卷积变换函数。沿着空间维度，将中间特征图分解为两个单独的特征张量，在两个卷积变换函数、两个 Sigmoid 激活函数的辅助下完成进一步计算，得到水平方向和垂直方向特征注意力[17]权值。

$$\begin{cases} \phi_1 = \delta'(\varepsilon'(\beta'''_1)) \\ \phi_2 = \delta''(\varepsilon''(\beta'''_2)) \end{cases} \quad (10)$$

式中，$\beta'''_1$、$\beta'''_2$ 表示两个拆解后形成的特征张量，$\phi_1$、$\phi_2$ 表示注意力权值，$\delta'$、$\delta''$ 表示两个卷积变换函数，$\varepsilon'$、$\varepsilon''$ 表示两个 Sigmoid 激活函数。利用公式（10）计算结果对输入特征图进行重新赋权，最终得到输出特征图为：

$$\bar{\beta}_v(u, p) = \beta'' R_u(u, p) \times \phi_{1v}(u) \times \phi_{2p}(u) \quad (11)$$

式中，$\bar{\beta}$ 表示最终输出特征图。经过主干网络和颈部网络的处理，可以得到智能驾驶车辆前方图像对应的特征图，图中蕴含了丰富的细节信息，是后续多目标检测的基础，将特征图导入头部网络的四个检测头中。每个检测头都会基于预设的锚框（Anchor）对特征图中的每个像素进行置信度计算和边界框回归[18]，从而预测出每个目标的位置、类别。

**1.3 基于点云 3D 投影生成多目标跟踪检测结果**

通过 YOLOv5 网络，得出智能驾驶车辆前方多目标检测结果后，为了能够对每个目标进行准确跟踪，需要融合环境激光扫描数据。利用激光扫描设备获取的车辆前方的实时点云数据，包含车辆前方环境的详细三维信息。运用点云 3D 投影原理[19]，推断出算法比较相邻帧之间目标在投影坐标系中位置变化的关联性，将 YOLOv5 算法检测出的目标，以激光束轮廓线的形式不断投射到三维环境中，以便实时跟踪目标位置变化。

点云 3D 投影需要依靠二维振镜扫描光学系统实现，在 X 轴和 Y 轴的检流计上，安装两个相互垂直的反射镜，借助反射镜的旋转动作，可以将激光光束按照预设的顺序投射到各个反射目标头上，以便对智能驾驶车辆前方环境进行多放线光束扫描[20]。多目标投影过程中，需要对每个反射目标头圆心的实际空间位置进行细致分析，依据反射目标头的位置信息，可以算出检流计所需达到的角度，利用这些角度信息以及反射目标头在世界坐标系中的理论坐标值，解算出具体的校准参数[21]，这些参数反映了投影坐标系与目标反射头以及投影

目标所共有的世界坐标系之间的转换关系[22]。一旦建立了这种转换关系，就能够轻松地将基于 YOLOv5 检测出的多目标坐标值从世界坐标系转换到投影坐标系中。这一过程中，点云 3D 投影系统内部几何关系如图 4 所示。

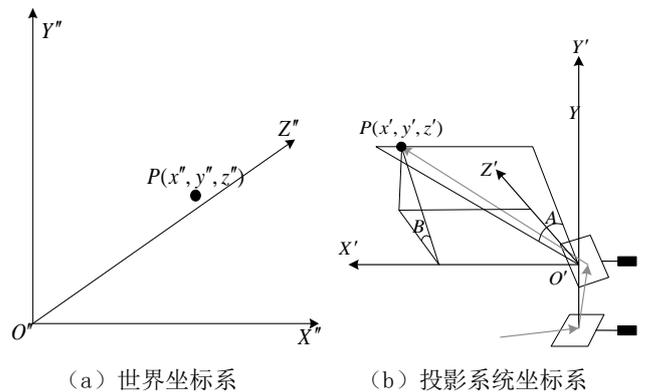

（a）世界坐标系 （b）投影系统坐标系

**图 4 点云 3D 投影系统内部几何关系**

图 4 中，$O''X''Y''Z''$ 表示世界坐标系，$O'X'Y'Z'$ 表示投影系统坐标系，$P$ 表示目标点，$(x'', y'', z'')$ 表示目标点在世界坐标系内的坐标值，$(x', y', z')$ 表示目标点在投影系统坐标系内的坐标值，$A$ 表示水平角，$B$ 表示俯仰角。明确点云 3D 投影系统内部几何关系后，通过公式（12）描述投影系统坐标系下目标点与水平角、目标点与俯仰角之间关系。

$$\begin{cases} x' = e \times \tan A + \varphi \times \tan A / \cos B \\ y' = \varphi \times \tan B \\ z' = \varphi \bar{\beta}_v(u, p) \end{cases} \quad (12)$$

式中，$e$ 表示两个反射镜间隔距离，$\varphi$ 表示目标点高度，$\tan$ 表示正切函数，$\cos$ 表示余弦函数。而世界坐标系与投影系统坐标系之间，目标点的坐标转换关系为：

$$\begin{bmatrix} x' \\ y' \\ z' \end{bmatrix} = \eta \begin{bmatrix} x'' \\ y'' \\ z'' \end{bmatrix} + T \quad (13)$$

式中，$\eta$ 表示旋转矩阵，$T$ 表示平移向量。从常规坐标转换算法中，选用具有实用性和稳定性优势的四元数法[23]，求解坐标转换所需的辅助参数，即可得到：

$$\eta = \begin{bmatrix} \tau_0^2 + \tau_1^2 - \tau_2^2 - \tau_3^2 & 2(\tau_1\tau_2 - \tau_0\tau_3) & 2(\tau_1\tau_3 - \tau_0\tau_2) \\ 2(\tau_1\tau_2 + \tau_0\tau_3) & \tau_0^2 - \tau_1^2 + \tau_2^2 - \tau_3^2 & 2(\tau_2\tau_3 - \tau_0\tau_1) \\ 2(\tau_1\tau_3 - \tau_0\tau_2) & 2(\tau_2\tau_3 + \tau_0\tau_1) & \tau_0^2 - \tau_1^2 - \tau_2^2 + \tau_3^2 \end{bmatrix} \quad (14)$$

式中，$\tau_0$ 表示四元数的实部，$\tau_1$、$\tau_2$、$\tau_3$ 表示四元数的虚部。明确不同坐标系之间的精确转换关系后，利用先进的激光扫描设备捕捉自动驾驶车辆前方环境 3D 点云数据[24]，这些数据包含了丰富的空间信息，是生成跟踪检测结果的关键。根据 3D 环境中点云数据分布特点，规划出一个投影平面，将连续帧图像上检测出的多目标不断投影到该平面上，即可直接跟踪每个运动目标的连续运动轨迹[25]，为后续智能驾驶车辆的行动提供依据。

## 2 实验分析

将 YOLOv5 网络和点云 3D 投影技术结合起来，设计出智能驾驶车辆前方多目标跟踪检测新方法后，需要对该方法进行实验分析，验证该方法在复杂交通环境下的目标跟踪能力。

实验分析的主要目的，可以从以下几个方面进行概括：首先，需要验证该融合方法在实际驾驶场景中的有效性。通过模拟真实道路条件，测试系统能否准确识别并跟踪前方的多个目标，包括但不限于车辆、行人及非机动车等，从而确保智能驾驶系统的安全性与可靠性。其次，在夜间、逆光、强光等复杂光照环境中，分别完成智能车辆前方多目标跟踪检测，观察这些条件下多目标检测精度与跟踪稳定性，评估该方法的综合性能，确保其在实际应用中能够稳定、准确地工作，为智能驾驶技术的发展提供坚实的技术支撑。

本文研究在智能驾驶车辆前方多目标跟踪检测方面考虑周全。实验场景设定在包含环岛、林荫路等多种典型驾驶环境的示范区内，模拟真实世界多样性，且挑选晴朗光照条件良好的天气在不同时间段测试。同时，涵盖直线行驶、环岛通过等多种驾驶场景，速度设置也各不相同。此外，采集大量数据建立训练集和测试集，样本量足够大以支持统计分析，还通过多种评估指标全面评估方法性能，确保研究结果的可靠性和有效性。

### 2.1 实验平台搭建

为了加强实验结果的说服力，实验场景设定在具有复杂道路特征的示范区内，该区域包含了环岛、林荫路、隧道等多种典型驾驶环境，以模拟真实世界的多样性。邀请六位风格各异的驾驶员（三名男性和三名女性）作为实验对象，他们均持有 C1 机动车驾驶证，并在试验前确保了充足的休息，以避免疲劳驾驶对实验结果产生影响。挑选了晴朗且光照条件良好的天气，在不同时间段进行实验测试，每位驾驶员将按照预设的路线和速度进行驾驶，路线设计涵盖了直线行驶、环岛通过、林荫路行驶、隧道穿越、S 弯和直角弯等多种驾驶场景。在直线行驶阶段，车辆将以 35km/h 的速度前进，以评估模型在稳定驾驶条件下的目标检测能力。通过环岛时，速度将降至 25km/h，以测试模型在复杂交通交汇点的表现。林荫路行驶速度设为 30km/h，旨在考察模型在树荫遮挡情况下的检测精度。进入隧道后，车辆将以 20km/h 的速度行驶，此时将特别关注模型在低光照条件下的鲁棒性。选择数据采集设备为双目视觉相机、激光扫描设备、激光跟踪仪等，计算设备为 NVIDIA A100 GPU，该设备具有强大的计算能力，支持 YOLOv5 算法的高效运行。软件工具有 YOLOv5 算法，该算法基于 PyTorch 深度学习框架，用于实时目标检测和分类。并通过点云 3D 投影技术，将点云数据投影到三维空间，增强目标跟踪的精度和鲁棒性。

在上述场景中完成实验操作之前，考虑基于 YOLOv5 和点云 3D 投影的智能驾驶车辆前方多目标跟踪检测方法工作要求，在智能驾驶车辆上安装双目视觉相机、激光扫描设备、智能激光 3D 投影系统等主要硬件设备，前两项设备主要负责在车辆行驶过程中采集环境数据。智能激光 3D 投影系统是由激光跟踪仪、激光 3D 投影设备和 GPS 测量系统组成，在实验过程中可以辅助点云 3D 投影技术的实施，从而保证车辆前方多目标跟踪检测实验的顺利开展。为了简化实验计算步骤，在智能激光 3D 投影系统中，设置了 3 个投影架标定点和 4 个承接部件基准点，通过一系列标定操作确定各点标定值如表 1 所示。

表 1 投影架与投影承接部件标定值

| 项目 | x/mm | y/mm | z/mm |
| --- | --- | --- | --- |
| 投影架标定点 1 | -260.44 | 312.64 | 12.64 |
| 投影架标定点 2 | -114.53 | 309.14 | -37.54 |
| 投影架标定点 3 | -21.19 | 306.44 | 18.65 |
| 承接部件基准点 1 | 1.36 | 503.59 | 56.46 |
| 承接部件基准点 2 | -204.97 | 501.24 | 507.84 |
| 承接部件基准点 3 | 4.36 | 506.36 | 494.16 |
| 承接部件基准点 4 | -207.86 | 502.67 | 10.02 |

在基础实验环境中，运用基于 YOLOv5 和点云 3D 投影的新方法、文献[5]提出的基于 3D 激光雷达点云处理的方法、文献[7]提出的基于改进 KCF 的方法、文献[8]提出的基于 DeepSort 的方法，分别完成智能驾驶车辆前方多目标跟踪检测实验。

### 2.2 数据集与评估指标

利用高精度卡口摄像机和激光雷达传感器，同步采集城市街道、高速公路、环岛、隧道等多种道路交通场景的视频流和点云数据，获取多目标跟踪检测所需的多样化数据。采集到的视频数据经过解码处理，可以提取出不同时间、地点和角度的图片帧，考虑到视频帧率较高，相邻帧间存在大量重复信息，通过比较图像的哈希值或特征相似度来识别并删除重复样本，减少数据集的冗余。应用 LabelImg 软件对去重后的图像数据进行标注，共确定了七个车辆驾驶场景目标类别，分别是轿车（car）、自行车（bicycle）、行人（people）、卡车（truck）、公交车（bus）、三轮车（tricycle）、摩托车（moto）。标注人员需要仔细标注每张图像中的目标，生成对应的 XML 文件，这些文件包含了目标的类别、位置（以边界框的形式给出）以及置信度等信息，为后续多目标跟

踪检测提供依据。由于 YOLOv5 算法无法处理空目标文件，为了保证数据集的完整性和一致性，编写了一个 Python 脚本，用于批量检查 XML 文件，并删除那些不包含任何标注目标的 XML 文件及其对应的 JPG 图片。

通过上述操作后，最终建立一个实验训练集和一个实验测试集，二者包含的图像数量分别为 30000 张、6000 张（考虑到数据清洗后的数量变化，实际数量略有不同）。为了满足 YOLOv5 算法对数据集格式的要求，将数据集转换为 VOC 格式，并进行了相应的目录结构调整。同时，还需要激光雷达采集的点云数据进行预处理，包括去噪、配准和投影到二维图像平面，以便与图像数据结合使用，增强多目标跟踪检测的能力。

为了对多目标跟踪检测方法的应用性能进行准确评估，采用了一个综合性强、涵盖多个维度的评价指标-多目标跟踪精度（MOTA）。MOTA 的计算过程相当复杂且精细，它综合考虑了四个关键要素：假阳性（FP）、假阴性（FN）、轨迹身份切换次数（IDS）以及真实框总数量（GT）。该指标不仅能够揭示算法在目标检测方面的优劣，还能够反映出算法在长时间跟踪过程中的稳定性和准确性，有助于推动多目标跟踪技术的不断进步和发展。

$$M = 1 - \frac{\sum_{o=1}^{n}(F_o + P_o + I_o)}{\sum_{o=1}^{n} G_o} \times 100\% \quad (16)$$

式中，$M$ 表示 MOTA（多目标跟踪精度）值，$n$、$o$ 分别表示总帧数和帧编号，$F$ 表示多目标跟踪结果的假阴性值（实际存在但在跟踪过程中未被检测出来的样本数量），$P$ 表示假阳性值（实际不存在但在跟踪过程中被错误检测出来的样本数量），$I$ 表示轨迹身份切换次数，$G$ 表示帧图像内真实框总数量。

**2.3 多目标识别结果**

YOLOv5 智能检测网络模型，多目标跟踪检测的重要环节，在实验过程中需要对模型进行训练，将其性能调整到最优。实际训练过程中，将网络超参数设置为表 2。

表 2 YOLOv5 网络参数

| 参数名称 | 数值 |
| --- | --- |
| 初始学习率 | 0.01 |
| 学习率动量 | 0.937 |
| 批次大小 | 32 |
| 最大迭代轮数 | 32200 |

初始学习率设置为 0.01 是通过网格搜索在 {0.1, 0.01, 0.001} 中选择，0.01 在训练初期能快速跳出局部最优，且避免梯度爆炸；学习率动量 0.937 是参考 SGD with Momentum 的默认配置，平衡收敛速度与稳定性。实验表明，动量值＞0.95 会导致损失曲线震荡；批次大小 32 受限于 GPU 显存（实验中使用的 NVIDIA A100 显存为 40GB），批次增大至 64 时显存不足，而 16 批次会导致梯度估计方差增大，降低收敛效率；最大迭代轮数 32200 由训练集规模（30000 张图像）和批次大小 32 计算得出（30000/32≈937.5 轮），额外增加 20% 迭代次数确保充分收敛。不同参数值对模型性能的影响：

学习率影响：当学习率过高（如 0.1）时，损失值在初期快速下降后陷入震荡，mAP 稳定在 92%；当学习率过低（如 0.001）时，收敛速度减慢，需增加 50% 迭代次数达到相同 mAP。

批次大小影响：16 批次时训练时间缩短 30%，但 mAP 下降 2.3%；64 批次时显存不足，需采用梯度累积策略，等效于 32 批次性能。

标定参数影响：标定点数量减少至 2 个投影架点，重投影误差增大 1.8 倍，导致跟踪精度下降 15%；激光跟踪仪采样频率从 10Hz 降至 5Hz；动态场景下跟踪延迟增加 3 帧，MOTA 值降低 8。在表 2 所示的参数条件下，应用训练集数据完成 300 次迭代运算，YOLOv5 网络模型的损失值和 mAP 值变化情况如图 5 所示。

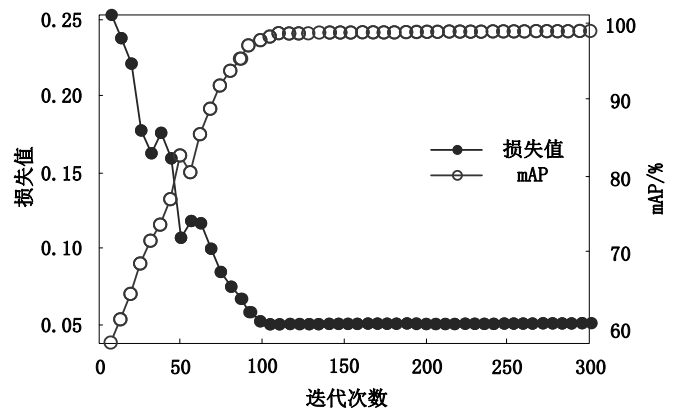

**图 5 YOLOv5 网络模型训练结果**

从图 5 可以看出，YOLOv5 网络模型随着不断训练，多目标检测结果的 mAP 值不断上升，而损失值则不断下降，这表明其工作性能正在不断优化。在损失值和 mAP 值变化呈现出稳定状态后，二者取值分别达到 0.05、99%，此时 YOLOv5 网络模型应用性能处于最优状态，将其应用到多目标跟踪检测过程中，可以实现对小目标的准确识别。

以四种不同驾驶场景对应的视觉图像为例，借助 YOLOv5 网络模型可以检测出每个场景中车辆前方多目标，最终呈现出的可视化检测结果如图 6 所示。

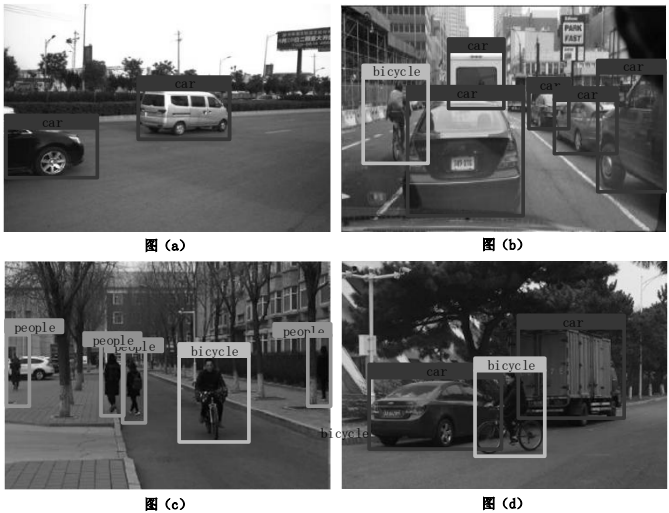

图(a)　　　　　　　　　图(b)

图(c)　　　　　　　　　图(d)

**图 6 车辆前方多目标检测结果**

从图 6 可以看出，无论是简单驾驶场景图像还是复杂驾驶场景图像，输入 YOLOv5 网络模型中均可以完成多目标检测，且最终检测结果中对于不同类型的目标，给出了不同标注结果，这表明了基于 YOLOv5 的多目标跟踪检测方法具有良好的目标检测性能。

**2.4 多目标跟踪结果**

以图 6（a）所示的公路场景为例，运用 YOLOv5 网络模型检测出智能驾驶车辆前方多目标后，借助点云 3D 投影技术可以对每个运动目标进行跟踪，最终得到图 7 所示的可视化跟踪结果。

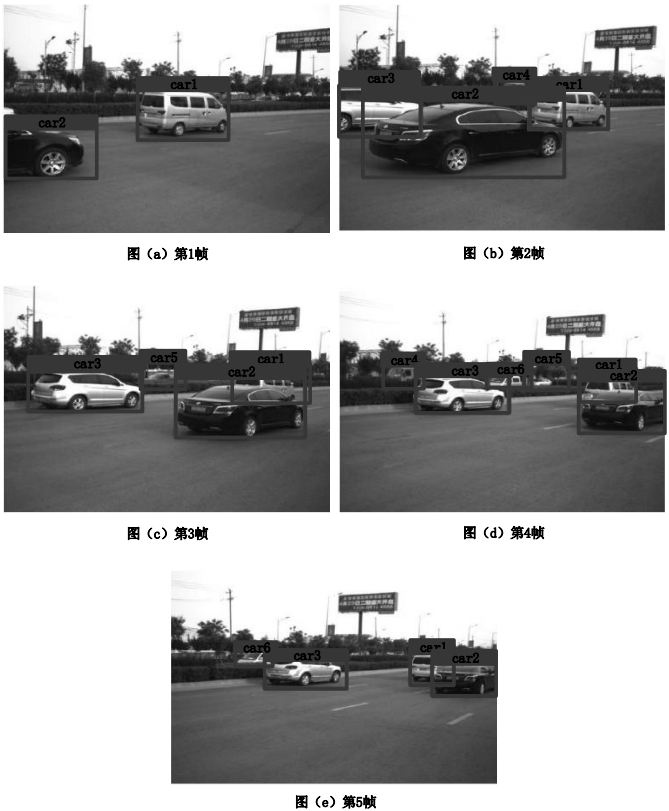

图（a）第1帧　　　　　　图（b）第2帧

图（c）第3帧　　　　　　图（d）第4帧

图（e）第5帧

**图 7 车辆前方多目标跟踪结果**

如图 7 所示，在连续 5 帧智能驾驶车辆前方环境图像中，编号为 car1 的车辆最初位于图像中间，随着不断运动最终只剩车尾。而编号为 car2 的车辆，在第 1 帧图像中只出现小部分车头，后续依次出现完整车辆，到第 5 帧只剩下车辆后半部分。除此之外，其余车辆在每帧图像中也得到了准确检测和标注。这一跟踪结果，证明了新设计方法在多目标跟踪检测过程中可以发挥优越性能。

**2.5 方法性能对比分析**

获取基于 YOLOv5 和点云 3D 投影的多目标跟踪检测结果后，运用其余三种文献提出方法，在同样的场景中完成多目标跟踪检测实验。设置待检测图像帧数逐步增加，从 500 帧逐渐过渡到 5500 帧，统计各方法输出的多目标跟踪检测结果，通过公式（16）推导出评估指标值，并生成图 8 所示的对比图像。

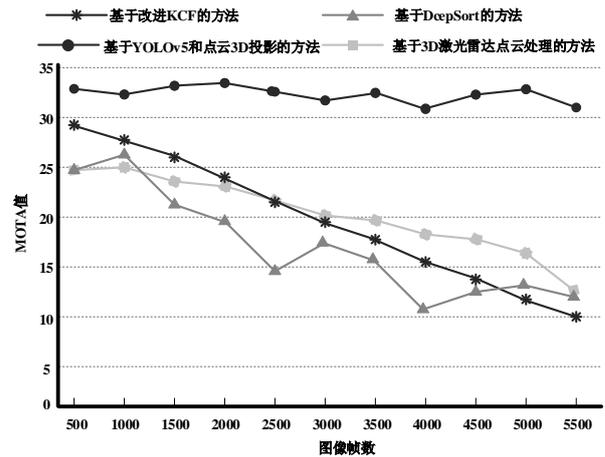

**图 8 各方法多目标跟踪检测结果 MOTA 值对比**

从图 8 给出的数据对比结果可以看出，三种文献方法在应对不断增加的图像帧数时，给出的多目标跟踪检测结果 MOTA 值，表现出明显的下降趋势，从 25 以上降低到 15 以下，这一变化反映出现有跟踪检测方法在处理高密度、快速变化的目标跟踪任务时，具有很大的局限性。

相比之下，基于 YOLOv5 和点云 3D 投影的创新方法展现出了截然不同的性能表现，在图像帧数不断增加的情况下，该方法生成的多目标跟踪检测结果 MOTA 值始终稳定在 30 以上。这主要因为 YOLOv5 作为一种先进的实时目标检测算法，以其高速和准确性著称，而结合点云 3D 投影技术，则进一步增强了模型对三维空间信息的理解和利用能力。这种组合不仅提升了目标检测的精确度，还有效解决了传统方法在复杂场景下的目标遮挡、深度信息缺失等问题。

通过对比可以看出，新研究跟踪检测方法在复杂驾驶环境中的鲁棒性和可靠性得到了极大增强。将其应用于智能驾驶系统，可以实现对周围环境的更精准感知，极大地提高了驾驶的安全性。

**3 结束语**

面向智能驾驶车辆前方多目标跟踪检测问题，提出基于 YOLOv5 和点云 3D 投影的新方法，实现了在复杂道路环境中对多个目标的精准跟踪与识别。该方法利用 YOLOv5 模型的高效检测能力，结合精心设计的锚框策略，能够快速定位并分类前方车辆、行人、障碍物等关

键目标，显著提升了跟踪的实时性和准确性。引入点云3D投影技术，将二维图像信息扩展至三维空间，不仅丰富了目标对象的几何特征，还有效解决了因遮挡、光照变化等导致的目标跟踪难题。通过精确计算点云数据的空间位置与分布，该方法能够在复杂场景下构建出更加真实、准确的目标模型，进一步增强了目标跟踪的稳定性和鲁棒性。综合来看，基于YOLOv5和点云3D投影的新型检测方法推广应用，为智能驾驶技术的发展注入了新的活力，标志着智能驾驶领域迈向了一个全新的发展阶段。

## 参考文献

[1]郭紫祎,张红娟,赵智博,等.基于路侧激光雷达的车辆目标跟踪与定位[J].测绘通报,2024,(12):84-89.
[2]王轩慧,吴颖,邵凯扬,等.基于改进YOLOv8s的自动驾驶多目标跟踪检测研究[J].汽车技术,2024,(12):1-7.
[3]张喜清,李进,陈殿民,等.基于模型预测控制的智能车辆轨迹跟踪仿真[J].计算机仿真,2024,41(10):114-120.
[4]郑策,董超,郑兵,等.面向方位历程交叉场景的多目标检测前跟踪方法[J].声学学报,2024,49(05):990-1004.
[5]武宏伟,吕东升,贾琳.一种用于3D激光雷达点云处理的多目标跟踪算法[J].信息与控制,2024,53(04):508-519.
[6]张瑶,卢焕章,王珏,等.短时记忆与CenterTrack的车辆多目标跟踪[J].中国图象图形学报,2023,28(10):3107-3122.
[7]刘毅,庞大为,田煜.基于改进KCF的多目标人员检测与动态跟踪方法[J].工矿自动化,2023,49(11):129-137.
[8]何维堃,彭育辉,黄炜,等.基于DeepSort的动态车辆多目标跟踪方法研究[J].汽车技术,2023,(11):27-33.
[9]王光明,宋亮,沈玥伶,等.基于目标检测和场景流估计联合优化的3D多目标跟踪[J].机器人,2024,46(05):554-561.
[10]任泽林,庞澜,王超,等.基于自编码器结构与改进Bytetrack的低光照行人检测及跟踪算法[J].应用光学,2024,45(03):616-629.
[11]朱奇光,商健,刘博,等.基于无人机航拍视频车辆多目标跟踪算法研究[J].计量学报,2024,45(12):1772-1779.
[12]卢锦,马令坤,吕春玲,等.基于代价参考粒子滤波器组的多目标检测前跟踪算法[J].自动化学报,2024,50(04):851-861.
[13]李志安,林道程,姜晓凤,等.基于改进YOLOv5算法和DeepSort算法的多目标检测和跟踪[J].济南大学学报(自然科学版),2024,38(05):556-563.
[14]王博,柴锐.基于改进YOLOV7的变规模网络重叠区域多目标跟踪方法[J].现代电子技术,2024,47(12):57-61.
[15]付小珊,胡乃平,秦建伟,等.概率扩充和改进OIM损失的多目标跟踪算法[J].计算机工程与设计,2024,45(07):2187-2194.
[16]蒲玲玲,杨柳.改进YOLOv5的多车辆目标实时检测及跟踪算法[J].科学技术与工程,2023,23(28):12159-12167.
[17]金佳男,吴延峰,史振宁,等.面向校园复杂环境的无人车前方多目标跟踪算法研究[J].现代制造工程,2023,(08):44-52.
[18]王学敏,于洪波,张翔宇,等.基于Hough变换检测前跟踪的水下多目标被动检测方法[J].兵工学报,2023,44(07):2114-2121.
[19]韩锟,彭晶莹.基于改进YOLOX与多级数据关联的行人多目标跟踪算法研究[J].铁道科学与工程学报,2024,21(01):94-105.
[20]陈曦,王昱程,曹宇,等.融合YOLO检测的孪生网络目标跟踪[J].武汉大学学报(工学版),2023,56(05):614-624.
[21]薄钧天,张嘉毫,王国宏,等.一种双重积累自反馈优化的三维多目标检测前跟踪算法[J].电子与信息学报,2024,46(09):3629-3636.
[22]姬张建,薛冰心.融合关系网络的Tracktor++多目标跟踪算法[J].山西大学学报(自然科学版),2023,46(05):1076-1084.
[23]余明骏,刁红军,凌兴宏.基于轨迹掩膜的在线多目标跟踪方法[J].山东大学学报(工学版),2023,53(02):61-69.
[24]周云,胡锦楠,赵瑜,等.基于卡尔曼滤波改进压缩感知算法的车辆目标跟踪[J].湖南大学学报(自然科学版),2023,50(01):11-21.
[25]吴开阳,秦文虎,云中华,等.基于激光雷达的三维多目标检测与跟踪[J].传感器与微系统,2023,42(01):122-125+130.

**作者简介：**

刘大勇，（1978），女，（汉）吉林长春人，硕士，高级工程师 主要研究方向：电子科学与技术专业

张清睿，（2004），男，（汉）山西长治人，本科生，主要研究方向：目标检测，yolo算法改进方向

孟泽阳，（2003），男，（汉）河南郑州人，本科生，主要研究方向：深度学习，目标检测方向

**联系方式：**

吉林省长春市朝阳区前进大街2699号吉林大学前卫南区